\documentclass{INTERSPEECH2023}
\usepackage{amsmath,graphicx}
\usepackage{amsfonts,amssymb}
\usepackage{booktabs}

\usepackage{algorithm}
\usepackage{algorithmic}
\usepackage{amssymb}
\usepackage{bbding}
\usepackage{amsmath}
\usepackage{multirow}
\usepackage{enumitem}

\usepackage{newfloat}
\usepackage{listings}

\interspeechcameraready

\usepackage{hyperref}
\hypersetup{
    colorlinks=true
    }
    
\newcommand\blfootnote[1]{%
  \begingroup
  \renewcommand\thefootnote{}\footnote{#1}%
  \addtocounter{footnote}{-1}%
  \endgroup
}

\title{MT4SSL: Boosting Self-Supervised Speech Representation Learning \\
by Integrating Multiple Targets}
\name{ 
Ziyang Ma\textsuperscript{1}, 
Zhisheng Zheng\textsuperscript{1}, 
Changli Tang\textsuperscript{2}, 
Yujin Wang\textsuperscript{2}, 
Xie Chen\textsuperscript{1}$^\ast$
}
\address{
$^1$ MoE Key Lab of Artificial Intelligence, AI Institute,  \\
X-LANCE Lab, Department of Computer Science and Engineering, \\ 
Shanghai Jiao Tong University, Shanghai, China  \\
$^2$ Department of Electronic Engineering, Tsinghua University, Beijing, China }
\email{\{zym.22, chenxie95\}@sjtu.edu.cn}

\begin{document}

\maketitle
 
\begin{abstract}
In this paper, we provide a new perspective on self-supervised speech models from how the training targets are obtained. 
We generalize the targets extractor into Offline Targets Extractor (Off-TE) and Online Targets Extractor (On-TE). 
Based on this, we propose a new multi-tasking learning framework for self-supervised learning, MT4SSL, which stands for Boosting Self-Supervised Speech Representation Learning by Integrating Multiple Targets. 
MT4SSL uses the K-means algorithm as an Off-TE and a teacher network without gradients as an On-TE, respectively. 
Our model outperforms previous SSL methods by nontrivial margins on the LibriSpeech benchmark, and is comparable to or even better than the best-performing models with fewer data. 
Furthermore, we find that using both Off-TE and On-TE results in better convergence in the pre-training phase. 
With both effectiveness and efficiency, we think doing multi-task learning on self-supervised speech models from our perspective is a promising trend. Code is available at \url{https://github.com/ddlBoJack/MT4SSL}. 
\end{abstract}
\noindent\textbf{Index Terms}: self-supervised learning, representation learning, multi-task learning, speech recognition


\section{Introduction}
\label{sec:Introduction}
\blfootnote{Corresponding author$^\ast$. }
Self-supervised learning (SSL) has achieved remarkable success in the field of representation learning, applied in computer vision~\cite{bao2021beit, he2022masked}, natural language processing~\cite{kenton2019bert, liu2019roberta}, as well as speech processing~\cite{baevski2020wav2vec, hsu2021hubert}. 
For speech representation learning, SSL methods are often used in the pre-training phase to obtain supervisory signals from massive unlabeled audio data. 

A core challenge for SSL is to obtain high-quality self-learning targets. 
Early works directly use input audio as training targets, either contrasting positive samples with negative ones~\cite{oord2018representation, schneider2019wav2vec}, or reconstructing the raw waves~\cite{chung2019unsupervised} and acoustic features~\cite{liu2021tera, ling2020deep}. 
Contemporary works explore many other ways to accomplish this goal, including quantization by quantizers~\cite{baevski2019vq, baevski2020wav2vec}, clustering by the K-means algorithm~\cite{hsu2021hubert, chen2022wavlm}, and generation by models~\cite{baevski2022data2vec}. 

In this work, we uniformly refer to the targets extraction module as the \textbf{Targets Extractor (TE)}.
Intuitively, we find that all TEs used in SSL models can be easily divided into two categories:
\begin{itemize}[leftmargin=*]
    \item Offline Targets Extractor (Off-TE). Off-TE is trained in advance, or is an off-the-shelf algorithm. Off-TE will not be updated in the self-supervised pre-training phase.
    \item Online Targets Extractor (On-TE). On-TE can be trained in advance, or randomly initialized. On-TE will be continuously updated during the pre-training process. 
\end{itemize}
It is obvious that targets from Off-TEs are more coarse-grained, and models using Off-TEs are easier to train. While On-TEs provide finer-and-finer-grained targets during training like curriculum learning does. Models with On-TEs might get better performance~\cite{baevski2022data2vec}. 
Another observation is that there is a clear complementarity on SUPERB\footnote{\url{https://superbbenchmark.org/leaderboard}} between HuBERT using Off-TE and data2vec using On-TE. 
Data2vec excels at content-related tasks, while HuBERT works better on speaker-related tasks. 
Therefore, we hope that both TEs can work together in a multi-task learning framework. 

We present \textbf{MT4SSL}, short for \textbf{M}ultiple \textbf{T}argets for \textbf{S}elf-\textbf{S}upervised \textbf{L}earning, to optimize the model with targets extracted from Offline Targets Extractor (Off-TE) and targets extracted from Online Targets Extractor (On-TE) simultaneously. 
Our design is similar to adopt the state-of-the-art models HuBERT~\cite{hsu2021hubert} and data2vec~\cite{baevski2022data2vec}. 
Actually, any two or more models using Off-TEs and On-TEs can be integrated into this framework. 
We conduct experiments on the LibriSpeech benchmark~\cite{panayotov2015librispeech}. 
With 360 hours of unlabeled data, Our model achieves an average of $10\%$ relative WER reduction over the best-performing HuBERT and data2vec on the 1-hour, 10-hour, and 100-hour fine-tuning subsets, and is comparable to or even better than wav2vec 2.0, HuBERT, and WavLM pre-trained with more data on the LibriSpeech 960h dataset. 
Furthermore, we find that MT4SSL can significantly speed up the convergence in the pre-training phase. The main contributions of this paper can be summarized as three-fold:
\begin{enumerate}[leftmargin=*]
\item We provide a new perspective on self-supervised speech models from how the self-training targets are obtained, and generalize the Targets Extractor (TE) into Offline Targets Extractor (Off-TE) and Online Targets Extractor (On-TE). 
\item We propose MT4SSL, a new multi-task learning framework that equips the model with both Off-TE and On-TE. We pre-train the model with targets extracted by both TEs and achieve better results than either alone on the LibriSpeech benchmark. 
\item We find that MT4SSL has good convergence. Compared with other models, relatively low WER on the speech recognition task can be obtained with only a few pre-training steps. We give a possible explanation for the good performance and convergence. 
We hope our findings could inspire researchers to develop more powerful self-supervised methods in the speech community. 
\end{enumerate}

\section{Related Works}
\label{sec:Related Works}
In this section, we introduce the progress of two key technologies in MT4SSL, including self-supervised learning (SSL) and multi-task learning (MTL).

\subsection{SSL on Speech Representation Learning}
Self-supervised pre-training followed by supervised fine-tuning becomes a mainstream approach for speech representation learning. 
This training paradigm has been shown to obtain universal representations on a wide variety of speech downstream tasks~\cite{yang2021superb}. 
Traditional categorization is to divide self-supervised methods into \textbf{1)} contrastive learning, \textbf{2)} predictive learning and \textbf{3)} generative learning based on the pretext tasks. 
Contrastive learning aims to distinguish positive samples from negative ones. 
CPC~\cite{oord2018representation} is the first successful representation learning approach for speech using contrastive learning, maximizing the mutual information between the input signal and the learned latent variables. 
Works that follow this paradigm include wav2vec~\cite{schneider2019wav2vec}, vq-wav2vec~\cite{baevski2019vq} and wav2vec 2.0~\cite{baevski2020wav2vec}. 
Predictive learning aims to predict the pre-clustered or model-generated targets with the input representations. 
HuBERT~\cite{hsu2021hubert} predicts the discrete targets clustered by the K-means algorithm of the masked regions with a BERT-like method. 
To learn better representations, WavLM~\cite{chen2022wavlm} polishes HuBERT by employing gated relative position bias and utterance mixing training strategy. ILS-SSL~\cite{wang2022improving} promotes HuBERT by adding an additional loss on the intermediate layers. PBERT~\cite{wang2022supervision} and HuBERT-AP~\cite{ren2022speech} refine HuBERT by refining target quality. 
Another predictive method is data2vec~\cite{baevski2022data2vec}, which generates high-quality targets for masked positions with a teacher model fed the same utterance. 
Generative learning aims to reconstruct the whole speech from latent variables with an auto-encoding model. 
These works focus on recovering discrete~\cite{chorowski2019unsupervised} or continuous~\cite{chen2019audio} speech signals using a variational autoencoder (VAE) or Seq2Seq autoencoder (SA) model to obtain representations of speech. 
Works that follow this paradigm include autoregressive models~\cite{chung2019unsupervised, chung2020vector} and non-autoregressive models~\cite{liu2021tera, ling2020deep}. 

\subsection{MTL for Speech Representation Learning}
There have been some works on MTL for speech representation learning. 
Early works enumerate as many self-supervised tasks as possible empirically to conduct MTL. 
PASE~\cite{pascual2019learning} and PASE+~\cite{ravanelli2020multi} solve many self-supervised tasks, such as waveform generation and prosody regression. 
Some concurrent works solve the problem with different pretext tasks. 
W2v-BERT~\cite{chung2021w2v} combines contrastive learning and mask language modeling to improve speech representation ability. 
They use the same targets extractor (quantizer) to obtain targets for different pretext tasks.
UniSpeech~\cite{wang2021unispeech} combines self-supervised learning and supervised learning to improve ASR performance. 
TESSP~\cite{yao2022tessp} and SpeechLM~\cite{zhang2022speechlm} enhance the representation ability of speech by introducing the mask language modeling task with paired or unpaired speech-text data.

\section{Method}
\label{sec:Method}
\begin{figure}[t]
    \centering
    \includegraphics[width=1.0\linewidth]{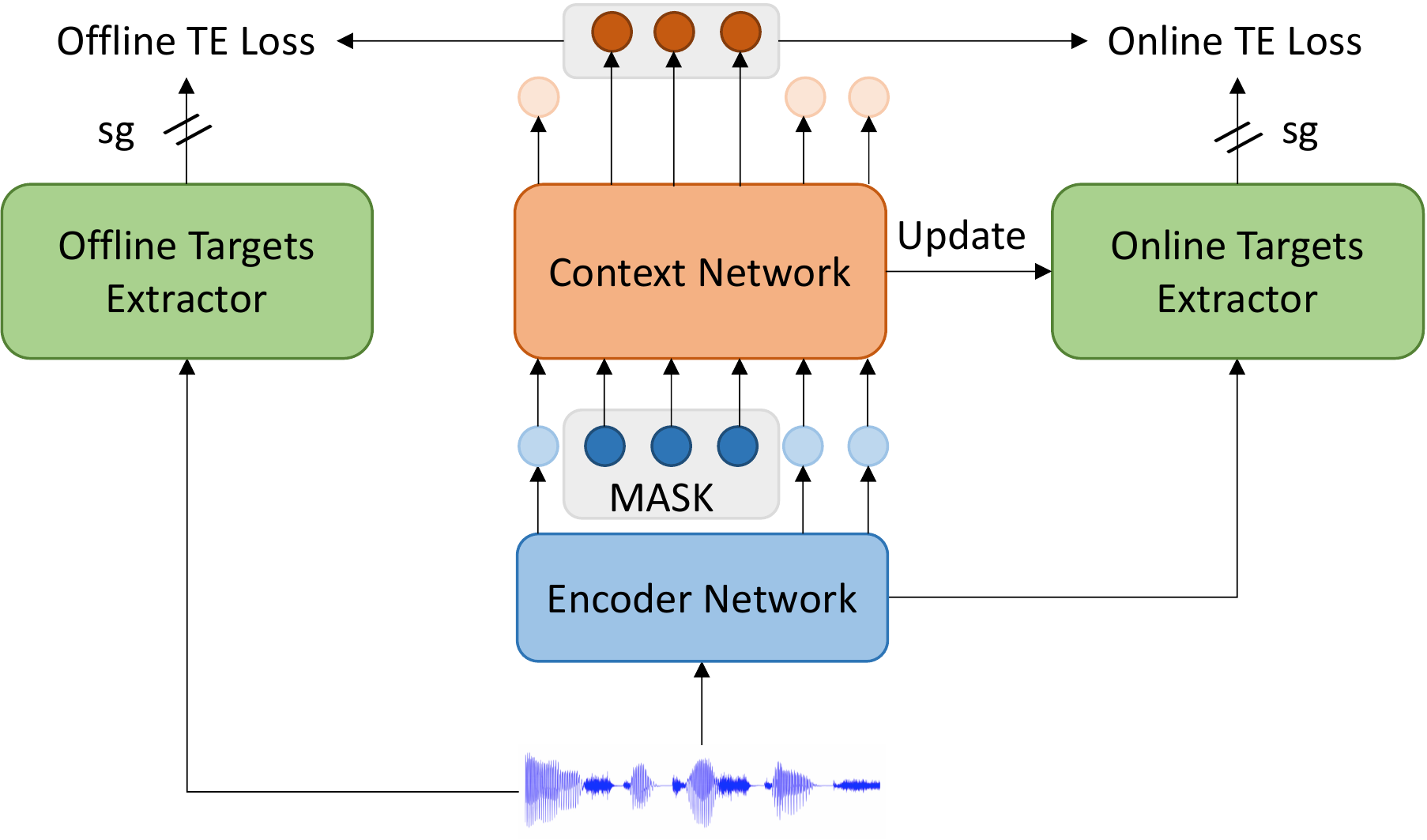}
    \caption{The overall framework of MT4SSL. The backbone of the model consists of an encoder network and a context network. The core of the framework is to optimize the model with multiple types of targets extracted from Off-TE and On-TE. sg means stop-gradient. }
    \label{fig:model}
\end{figure}

The overall architecture of \textit{MT4SSL} is shown in Figure~\ref{fig:model}. 
For the backbone network, we simply use the same architecture as mainstream baseline models~\cite{baevski2020wav2vec, hsu2021hubert, baevski2022data2vec}, which contains an \textit{encoder network} and a \textit{context network}. 
For the Target Extractor, we use the K-means algorithm as an \textit{Off-TE} as HuBERT does, and use a teacher network without gradients as an \textit{On-TE} as data2vec does.  
\subsection{Backbone Network}
\label{fig:Model Architecture}

The encoder network is a 7-layer 1-D convolutional neural network with kernel sizes $(5, 2, 2, 2, 2, 2, 2)$ and strides $(10, 3, 3, 3, 3, 2, 2)$. 
Given the raw audio input $\mathcal{X}$ at a 16000 Hz sample rate, we downsample the audio with the encoder network denoted with $f: \mathcal{X} \mapsto \mathcal{H}$. The output representations $\mathcal{H}$ are 50 Hz with dimension 512.
Then we apply a linear projection for dimension transformation from 512 to 768, followed by the mask matrix to construct the input of the context network, denoted with $m: \mathcal{H} \mapsto \mathcal{\tilde{H}}$.
The context network is a 12-layer standard Transformer with learnable convolutional positional encoding. Each Transformer block is set to 768 model dimension, 3072 inner dimension, and 12 attention heads.
The context network can be denoted with $g: \mathcal{\tilde{H}} \mapsto \mathcal{Z}$.
The final output $\mathcal{Z}$ of the backbone is used for classification and regression, which will be detailed in Section~\ref{sec:Discrete Targets} and Section~\ref{sec:Continuous Targets}.

\subsection{Offline Targets Extractor}
\label{sec:Discrete Targets}
The model parameters of Offline Targets Extractor (Off-TE) do not update during the pre-training phase. 
Here we use the K-means algorithm as the Off-TE.
We train a model with the K-means algorithm before pre-training.
The model transforms speech features into $C$ clusters. 
Thus we use the indices of cluster centers to represent each speech token. 
Suppose $X$ is a raw audio utterance sampled from $\mathcal{X}$, the offline targets can be obtained by:
\begin{equation}
Y^{f} = \mathbf{TE_{f}}(X),
\end{equation}
where $\mathbf{TE_{f}(\cdot)}$ is the extracting operation, and $Y^f = [y_1^f, \cdots, y_T^f]^\mathrm{T}$ is the self-training offline targets and $y_{t}^f \in \mathbb{R}^C$ in a one-hot form. 
Suppose $Z = [z_1, \cdots, z_T]^\mathrm{T}$ is a masked version obtained through the backbone network from the same raw audio $X$. $M \subset [T]$ denote the masked indices, and $z_t$ is replaced with a mask token if $t \in M$. 
We use a projection layer for dimension transformation, which can be written as:
\begin{equation}
Z^f = \mathbf{W}^f Z,
\end{equation}
where $Z^f = [z_1^f, \cdots, z_T^f]^\mathrm{T}$ and $z_{t}^f \in \mathbb{R}^C$. The offline TE loss is defined as:
\begin{equation}
\label{eq:discrete loss}
\mathcal{L}^f=\mathbf{CE}(Z^f, Y^f),
\end{equation}
where $\mathbf{CE(\cdot)}$ computes the cross entropy loss between sources and targets. 

\subsection{Online Targets Extractor}
\label{sec:Continuous Targets}
The model parameters of Online Targets Extractor (On-TE) update continuously during the pre-training phase. 
We use a teacher network without gradients as the On-TE to obtain online targets. This process can be viewed as a special kind of knowledge distillation \cite{grill2020bootstrap}, or Noisy Student Training (NST) \cite{park2020improved}. 
Suppose $H$ is the hidden representation sampled from $\mathcal{H}$, which is obtained from $\mathcal{X}$ followed by convolutional subsampling, the online targets can be obtained by:
\begin{equation}
Y^n = \mathbf{TE_n}(H),
\end{equation}
where $\mathbf{TE_n}(\cdot)$ is the extracting operation, and $Y^n = [y_1^n, \cdots, y_T^n]^\mathrm{T}$ is the self-training targets. 
As with offline speech tokens, we use a projection layer for dimension transformation, which can be written as:
\begin{equation}
Z^n = \mathbf{W}^n Z,
\end{equation}
where $Z^n$ has the same dimension as $Y^n$.  
The online TE loss is defined as: 
\begin{equation}
\label{eq:continuous loss}
\mathcal{L}^n=\mathbf{MSE}(Z^n, Y^n),
\end{equation}
where $\mathbf{MSE(\cdot)}$ measures the mean squared error between sources and targets. 
The parameters of the teacher network $\Delta$ are initialized with the parameters of the backbone network $\theta$, and the parameters of the teacher network are updated with exponentially moving average (EMA)~\cite{grill2020bootstrap} within each mini-batch, donated as:
\begin{equation}
\Delta = \tau \Delta+(1-\tau) \theta,
\end{equation}
where $\tau$ is a parameter that increases linearly during pre-training. 

\subsection{Loss Function}
\label{sec:Loss Function}

The core of our model is to integrate multiple targets, thereby enhancing the representational ability of self-supervised learning. The proposed objective is formulated as follows by using both offline TE loss in Eq.~\ref{eq:discrete loss} and online TE loss in Eq.~\ref{eq:continuous loss}:
\begin{equation}
\mathcal{L}=\mathcal{L}^f + \alpha\mathcal{L}^n,
\end{equation}
with a tunable parameter $\alpha$.
Note that we only compute loss on the masked parts of the utterance.

\section{Experiments}
\label{sec:Experiments}
\subsection{Dataset}
For unsupervised pre-training, we use LibriSpeech~\cite{panayotov2015librispeech} corpus with 360-hour unlabeled data (train-clean-360). For supervised fine-tuning,  1-hour, 10-hour splits from Libri-light~\cite{librilight} corpus and 100-hour from LibriSpeech corpus are considered. We conduct model evaluation according to the mainstream test sets: dev-clean/other and test-clean/other from the LibriSpeech corpus.
\subsection{Setup}
\label{sec:Setup}
Our MT4SSL model can be considered as a simplification and fusion of the HuBERT model and the data2vec model, so we maximized the inheritance of the hyperparameters of both to demonstrate the effectiveness of the model. 
Given the limited computing resources, we simply choose some empirical configuration for the training of MT4SSL model without conducting extensive hyperparameter search. 

\begin{table}[H]
\begin{center}
\begin{small}
\resizebox{\linewidth}{!}{
\begin{tabular}{l|c|cccc}
\hline
\hline
\multirow{3}*{Model}& \multirow{3}*{\shortstack{Language\\Model}}  & \multicolumn{4}{c}{WER$\%$($\downarrow$)}  \\
&  & \multicolumn{2}{c}{dev} & \multicolumn{2}{c}{test} \\ 
\cline{3-6}
&  & clean & other & clean & other \\ 
\hline
\hline
\multicolumn{6}{l}{\textit{1-hour Labeled Data}} \\
\hline
HuBERT & \multirow{3}*{None} & 29.4 & 37.1 & 30.0 & 37.8 \\ 
data2vec  &  & 24.1 & 32.6 & 24.2 & 33.5  \\ 
\textbf{MT4SSL}  &  & \textbf{19.7} & \textbf{27.1} & \textbf{20.1} & \textbf{27.6} \\ 
\hline
HuBERT  & \multirow{3}*{$4$-gram} &  7.2 & 16.5 & 7.9 & 17.3 \\ 
data2vec  &  &  6.8 & 15.8 & 7.2 & 17.0 \\ 
\textbf{MT4SSL}  &  &  \textbf{5.5} & \textbf{13.1} & \textbf{5.9} & \textbf{13.9} \\ 
\hline
\hline
\multicolumn{6}{l}{\textit{10-hour Labeled Data}} \\
\hline
HuBERT & \multirow{3}*{None} & 11.4 & 21.3 & 11.6 & 22.5  \\ 
data2vec &  & 10.8 & 20.9 & 10.7 & 21.9   \\ 
\textbf{MT4SSL} &  & \textbf{9.4} & \textbf{18.0} & \textbf{9.3} & \textbf{18.5} \\
\hline
HuBERT & \multirow{3}*{$4$-gram} & 4.9 & 12.7 & 5.3 & 13.4 \\ 
data2vec&  & 5.1 & 13.8 & 5.5 & 14.5  \\ 
\textbf{MT4SSL} &  &  \textbf{4.0} & \textbf{11.4} & \textbf{4.5} & \textbf{11.9} \\
\hline
\hline
\multicolumn{6}{l}{\textit{100-hour Labeled Data}} \\
\hline
HuBERT & \multirow{3}*{None} & 5.8 & 15.5 & 6.0 & 15.3  \\ 
data2vec &  & 5.5  & 15.7 & 5.6 & 16.0 \\ 
\textbf{MT4SSL} &  & \textbf{5.1} & \textbf{14.3} & \textbf{5.1} & \textbf{14.3} \\ 
\hline
HuBERT & \multirow{3}*{$4$-gram} & 2.9 & 9.6 & 3.5 & 9.9 \\ 
data2vec &  & 3.0 & 10.6 & 3.6 &11.2  \\ 
\textbf{MT4SSL} &  & \textbf{2.7} & \textbf{9.6} & \textbf{3.4} & \textbf{9.6} \\ 
\hline
\hline
\end{tabular}
}
\caption{Word error rate (WER) on LibriSpeech corpus. We compare the performance on four subsets (dev-clean, dev-other, test-clean, test-other) with ($4$-gram) and without (None) language model pre-trained on 360 hours of unlabeled data (train-clean-360) and fine-tuned on different amounts of labeled data ($1$h, $10$h, $100$h). }
\label{tab:Results}
\end{small}
\end{center}
\end{table}
\vspace{-1cm}

\begin{table}[H]
\begin{center}
\begin{small}
\resizebox{\linewidth}{!}{
\begin{tabular}{l|c|cccc}
\hline
\hline
\multirow{3}*{Model} & \multirow{3}*{\shortstack{Unlabeled\\Data}}  & \multicolumn{4}{c}{WER$\%$($\downarrow$)} \\
&  & \multicolumn{2}{c}{dev} & \multicolumn{2}{c}{test} \\ 
\cline{3-6}
&  & clean & other & clean & other \\ 
\hline
\hline
\multicolumn{6}{l}{\textit{1-hour Labeled Data}} \\
\hline
wav2vec 2.0\cite{baevski2020wav2vec} & LS-960 & 24.1 & 29.6 & 24.5 & 29.7 \\ 
HuBERT\cite{hsu2021hubert} & LS-960 & 24.3 & 30.2 & 20.9 & \textbf{27.5} \\ 
WavLM\cite{chen2022wavlm} & LS-960 & - & - & 24.5 & 29.2 \\ 
\textbf{MT4SSL} &\textbf{LS-360} & \textbf{19.7} & \textbf{27.1} & \textbf{20.1} & 27.6 \\ 
\hline
\hline
\multicolumn{6}{l}{\textit{10-hour Labeled Data}} \\
\hline
wav2vec 2.0\cite{baevski2020wav2vec} & LS-960 & 10.9 & 17.4 & 11.1 & 17.6 \\ 
HuBERT\cite{hsu2021hubert} & LS-960 & 12.0 & 18.1 & 10.1 & 16.8 \\ 
WavLM\cite{chen2022wavlm} & LS-960 & - & - & 9.8 & \textbf{16.0} \\ 
\textbf{MT4SSL} & \textbf{LS-360} & \textbf{9.4} & \textbf{18.0} & \textbf{9.3} & 18.5 \\
\hline
\hline
\multicolumn{6}{l}{\textit{100-hour Labeled Data}} \\
\hline
wav2vec 2.0\cite{baevski2020wav2vec} & LS-960 & 6.1 & 13.5 & 6.1 & 13.3 \\ 
HuBERT\cite{hsu2021hubert} & LS-960 & 5.5 & \textbf{13.0} & 6.3 & 13.2 \\ 
WavLM\cite{chen2022wavlm} & LS-960 & - & - & 5.7 & \textbf{12.0} \\ 
\textbf{MT4SSL} & \textbf{LS-360} & \textbf{5.1} & 14.3 & \textbf{5.1} & 14.3\\
\hline
\hline
\end{tabular}
}
\caption{Word error rate (WER) on LibriSpeech corpus. The results for wav2vec 2.0 and WavLM are obtained from their papers. The results for HuBERT are obtained by fine-tuning their public released model~\protect\footnotemark[2]. All results are reported without the language model. }
\label{tab:960}
\end{small}
\end{center}
\end{table}
\footnotetext[2]{\url{https://github.com/facebookresearch/fairseq/tree/main/examples/hubert}}
\vspace{-1cm}

\textbf{Pre-Training}.
In the pre-training phase, we train the model with 360 hours LibriSpeech unlabeled data. The training is conducted on NVIDIA GeForce RTX 3090 GPUs, and we simulate 16 GPUs by using $k$ GPUs and setting update frequency with $16/k$. $k$ is set to $4$ in this paper. 
For the mask strategy, each time step has a probability of $p = 0.065$ to be the starting index and the subsequent $l = 10$ time-steps are masked. This results in the mask embedding covering $49\%$ of all tokens on average. 
For the optimizing strategy, we use Adam~\cite{kingma2015adam} with a learning rate of $0.0005$ and a weight decay of $0.01$. We train MT4SSL for $800$ epoch, with $[3\%, 90\%, 7\%]$ proportion of warm-up, hold-on, and linearly decay. The hyperparameter  $\alpha$ that controls the loss weight is set to $1$, which means the two losses have the same weight. 

For the offline targets, we obtain features from HuBERT model and train an Off-TE before pre-training.
Concretely, the targets are obtained by running the K-means clustering with $500$ clusters on the $6$-th transformer layer output of the first iteration HuBERT model.

For the online targets, we use the average of the top $8$ blocks of the transformer layer outputs from the teacher network as data2vec model designs. 
For the parameter update of the teacher model, we apply a linearly increasing strategy for $\tau$ from $\tau_s=0.99$ to $\tau_e=0.999$ for the first $7.5\%$ training steps. The parameter $\tau$ is kept constant for the remainder of training steps. 

\textbf{Fine-Tuning}.
In the fine-tuning phase, we use Connectionist Temporal Classiﬁcation (CTC)~\cite{graves2006connectionist} loss to keep consistent with the baseline models. The hyper-parameters of the fine-tuning stage are still kept consistent with the mainstream models~\cite{baevski2020wav2vec,hsu2021hubert,baevski2022data2vec}.

\subsection{Results}

Tabel~\ref{tab:Results} shows results of the MT4SSL on the LibriSpeech benchmark compared to other state-of-the-art models. 
The models are pre-trained on LibriSpeech $360$ hours dataset (train-clean-$360$), and fine-tuned on Libri-light $1$-hour, $10$-hour and LibriSpeech $100$-hour subsets. 
We compare the performance on dev-clean/other and test/other with and without the language model, respectively.
We adopt the $4$-gram language model trained on the official LibriSpeech language modeling data. 
Given 10 hours of labeled data, MT4SSL can achieve $13.0\%$ (dev-clean), $13.9\%$ (dev-other), $13.1\%$ (test-clean) and $15.5\%$ (test-other) relative WER reduction over the best-performing model without a language model, and $18.4\%$ (dev-clean), $10.2\%$ (dev-other), $15.1\%$ (test-clean) and $11.2\%$ (test-other) relative WER reduction with a $4$-gram language model. 
For the fine-tuning on 1-hour and 100-hour labeled data, the improvements are consistent for the MT4SSL over other models. 

Table~\ref{tab:960} shows results of MT4SSL trained with 360 hours of audio data, compared to the state-of-the-art models trained with 960 hours. 
We compare results of MT4SSL with wav2vec 2.0 and WavLM from their papers, and HuBERT from their public released page. 
Despite using less data, our model is comparable to or even better than the state-of-the-art models. The performance on noisy subsets is less competitive, and one possible explanation is that our pre-training data (train-clean-360) only consists of clean audio.

\begin{figure}[ht]
    \centering
    \includegraphics[width=1.0\linewidth]{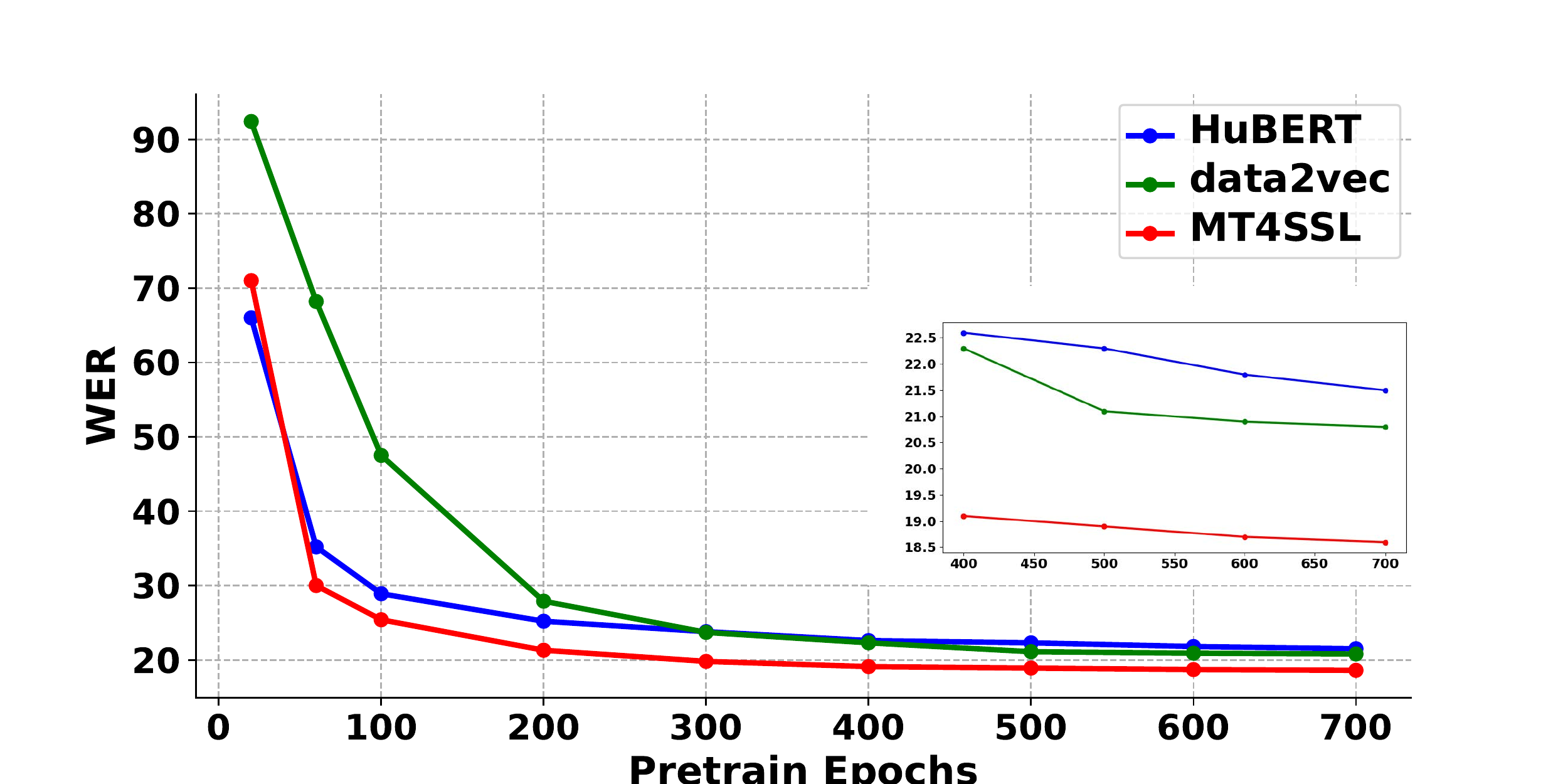}
    \caption{Word error rate (WER) along with pre-training epochs among different SSL models. The models are fine-tuned on $10$-hour labeled data and evaluated on LibriSpeech dev-other dataset without adpoting the language model.}
    \label{fig:pretrainSteps-wer}
\end{figure}

\subsection{Convergence Analysis}

In this section, we analyze our MT4SSL in terms of model convergence quantitatively and qualitatively. 
All experiments are carried out with the following configuration: all the models are pre-trained on 360 hours of LibriSpeech unlabeled data, fine-tuned on 10 hours of Libri-light labeled data, and evaluated on dev-other subset of LibriSpeech corpus.

We find that MT4SSL not only achieves amazing results on the benchmark, but also exhibits good convergence. 
As shown in Figure~\ref{fig:pretrainSteps-wer}, we plot the WER trend with respect to the number of pre-training epochs. 
By comparing MT4SSL with data2vec and HuBERT, it can be seen that HuBERT which utilizes offline targets has better convergence than data2vec which utilizes online targets. 
However, data2vec achieves better performance than HuBERT when fully trained. 
MT4SSL combines their advantages and converges to a relatively low WER quickly. 

One possible explanation is that the fixed offline targets are less difficult to learn for the model to learn than the ever-changing online targets. Hence, the model which adopts offline targets converges faster. However, the online targets have finer granularity than the offline targets, so the model which uses online targets has better representation capabilities. 
The learning of the two targets may not be conflicting but cooperative, so MT4SSL can achieve both efficiency and effectiveness.

\section{Discussion}
\label{sec:iscussion}
In this work, a preliminary attempt to combine Off-TE using the K-means algorithm and On-TE using a randomly initialized teacher has yielded good results. There are several interesting aspects to explore:
\begin{itemize}[leftmargin=*]
\item Will there be On-TEs and Off-TEs that cooperate better?
\item Will using different targets extractors at different stages of pre-training improve efficiency, effectiveness, or both?
\item Is MTL in SSL a better way to obtain universal features in various speech-related downstream tasks?  
\end{itemize}
We will research the above problems in the future. 

\section{Conclusion}
\label{sec:conclusion}
In this paper, we present a new framework for speech-based self-supervised learning, which is named MT4SSL, by simultaneously optimizing the model with offline targets and online targets, without caring about specific pretext tasks. 
The proposed method improves both performance and convergence upon the state-of-the-art models. 

\section{Acknowledgements}
This work was supported by the National Natural Science Foundation of China  (No. 62206171), the International Cooperation Project of PCL, and Alibaba Group through Alibaba Innovative Research Program.

\bibliographystyle{IEEEtran}
\bibliography{mybib}

\end{document}